\documentclass{bmcart}

\usepackage{amsthm,amsmath}
\RequirePackage{hyperref}
\usepackage[utf8]{inputenc} 
\usepackage{booktabs}
\usepackage{multirow}
\usepackage{colortbl}
\usepackage{graphicx}
\usepackage{subfig}

\startlocaldefs
\endlocaldefs

\begin{document}

\begin{frontmatter}

\begin{fmbox}
\dochead{Research}


\title{Developing a Portable Natural Language Processing Based Phenotyping System}


\author[
   addressref={aff1},                   
   noteref={n1},                        
   email={himanshu@uic.edu}             
]{\inits{HS}\fnm{Himanshu Sharma}}
\author[
   addressref={aff2},
   noteref={n1},
   email={chengsheng.mao@northwestern.edu}
]{\inits{CM}\fnm{Chengsheng Mao}}
\author[
   addressref={aff2},
   email={yizhenzhang2018@u.northwestern.edu}
]{\inits{YZ}\fnm{Yizhen Zhang}}
\author[
   addressref={aff1},
   email={hvatan2@uic.edu}
]{\inits{HV}\fnm{Haleh Vatani}}
\author[
   addressref={aff2},
   email={liang.yao@northwestern.edu}
]{\inits{LY}\fnm{Liang Yao}}
\author[
   addressref={aff2},
   email={yizhen.zhong@northwestern.edu}
]{\inits{YZ}\fnm{Yizhen Zhong}}
\author[
   addressref={aff2},
   email={luke.rasmussen@northwestern.edu}
]{\inits{LR}\fnm{Luke Rasmussen}}
\author[
   addressref={aff3},
   email={Jiang.Guoqian@mayo.edu}
]{\inits{GJ}\fnm{Guoqian Jiang}}
\author[
   addressref={aff4},
   email={jyp2001@med.cornell.edu}
]{\inits{JP}\fnm{Jyotishman Pathak}}
\author[
   addressref={aff2},
   corref={aff2},                       
   email={yuan.luo@northwestern.edu}
]{\inits{YL}\fnm{Yuan Luo}}
\address[id=aff1]{
  \orgname{Cyberinfrastructure, University of Illinois at Chicago}, 
  \city{Chicago},                              
  \cny{USA}                                    
}
\address[id=aff2]{%
  \orgname{Department of Preventive Medicine, Feinberg School of Medicine, Northwestern University},
  \street{750 N Lakeshore Dr.},
  \postcode{60611}
  \city{Chicago},
  \cny{USA}
}
\address[id=aff3]{
  \orgname{Biomedical Informatics, Mayo Clinic}, 
  \city{Rochester},                              
  \cny{USA}                                    
}
\address[id=aff4]{
  \orgname{Health Informatics, Weill Cornell Medicine}, 
  \city{New York},                              
  \cny{USA}                                    
}
\begin{artnotes}
\note[id=n1]{Equal contributor} 
\end{artnotes}
\end{fmbox}

\begin{abstractbox}
\begin{abstract} 
This paper presents a portable phenotyping system that is capable of integrating both rule-based and statistical machine learning based approaches. Our system utilizes UMLS to extract clinically relevant features from the unstructured text and then facilitates portability across different institutions and data systems by incorporating OHDSI’s OMOP Common Data Model (CDM) to standardize necessary data elements. Our system can also store the key components of rule-based systems (e.g., regular expression matches) in the format of OMOP CDM, thus enabling the reuse, adaptation and extension of many existing rule-based clinical NLP systems. We experimented with our system on the corpus from i2b2’s Obesity Challenge as a pilot study. Our system facilitates portable phenotyping of obesity and its 15 comorbidities based on the unstructured patient discharge summaries, while achieving a performance that often ranked among the top 10 of the challenge participants. This standardization enables a consistent application of numerous rule-based and machine learning based classification techniques downstream. 
\end{abstract}


\begin{keyword}
\kwd{NLP}
\kwd{Portability}
\kwd{Machine Learning}
\kwd{Obesity}
\kwd{i2b2}
\end{keyword}


\end{abstractbox}
%

\end{frontmatter}



\section*{INTRODUCTION}
The Electronic Health Record (EHR) is often described as “a longitudinal electronic record of patient health information generated by one or more encounters in any care delivery setting. Included in this information are patient demographics, progress notes, problems, medications, vital signs, past medical history, immunizations, laboratory data and radiology reports.”\cite{himss}  As medical care becomes more data-driven and evidence-based, these EHRs become essential sources of health information necessary for decision-making in all aspects of patient assessment, phenotyping, diagnosis, and treatment. 

These EHRs contain both a) structured data such as orders, medications, labs, diagnosis codes and unstructured data such as textual clinical progress notes, radiology and pathology reports. While structured data may not require significant preprocessing to derive knowledge, Natural Language Processing (NLP) techniques are commonly used to analyze unstructured data. This unstructured data can feed into a variety of secondary analysis such as clinical decision support, evidence-based practice and research, and computational phenotyping for patient cohort identification \cite{sarmiento2016improving}. Additionally, manual labeling of a large volume of unstructured data by the experts can be very time-consuming and impractical when used across multiple data sources. Automated information extraction from unstructured data through NLP provides a more efficient and sustainable alternative to the manual approach \cite{sarmiento2016improving}. 

As summarized in a 2013 review by Shivade et al. \cite{shivade2013review}, early computational phenotyping studies were often formulated as supervised learning problems wherein a predefined phenotype is provided, and the task is to construct a patient cohort matching the definition’s criteria. Unstructured clinical narratives may summarize patients’ medical history, diagnoses, medications, immunizations, allergies, radiology images, and laboratory test results, in the form of progress notes, discharge reports etc. and provide a valuable resource for computational phenotyping \cite{shickel2017deep}. While we refer the readers to reviews such as \cite{shivade2013review} for more details on phenotyping methods, we point out that information heterogeneity in clinical narratives asks for development of portable phenotyping algorithms. Boland et al. \cite{boland2013defining} highlighted the heterogeneity apparent in clinical narratives due to the variance in physicians’ expertise and behaviors, and institutional environments and setups. Studies have applied Unified Medical Language System (UMLS) or other external controlled vocabularies to recognize the various expressions of the same medical concept and standard UMLS annotations are generally considered a must for portable phenotyping \cite{hersh2000assessing,passos2009wordnet}. 

Our main aim was to introduce portability to the ongoing research efforts on NLP-driven phenotyping of unstructured clinical records. To this end, we leveraged a well-defined phenotyping problem, i2b2 Obesity Challenge, to perform a pilot study and introduced new steps to this multi-class and class-unbalanced classification problem for portability. We extracted structured information from 1249 patient textual discharge summaries by parsing each record through a context-aware parser (MetaMap) and mapped all of the extracted features to UMLS’s Concept Unique Identifiers (CUIs). MetaMap’s output was then stored in a MySQL database using the schemas defined in the Observational Medical Outcomes Partnership (OMOP) Common Data Model (CDM), a data standardization model championed by the Observational Health Data Sciences and Informatics (OHDSI) collaborative. 

We recognize the usefulness of existing rule-based (e.g., RegEx-driven) NLP systems and enable our system to introduce/improve their portability by storing key components of rule-based NLP systems as annotations using the format defined in the OMOP CDM. We explore the tradeoff between phenotyping accuracy and portability, which has been largely ignored but of critical importance. We evaluated a combination of rule-based (RegEx-driven) and machine learning approaches to assess the trade-off through an iterative manner for obesity and its 15 comorbidities. We ran four types of machine learning algorithms on our dataset, and conducted multiple iterations of optimizations for a balanced trade-off between classification performance and portability. In particular, Decision Tree resulted in the best performance with the F-Micro score for intuitive classification at 0.9339 and textual classification at 0.9546 and the F-Macro score for intuitive classification at 0.6509 and textual classification at 0.7855.

\section*{SYSTEM DESCRIPTION}
Our portable NLP system is based on sequential activities that form an NLP pipeline with six major components: a) Data Preparation and Environmental Setup, b) Section and Boundary Detection, c) Annotation Feature Extraction and Mapping, d) Regular Expression matches as Annotations, e) Classification \& f) Performance Tuning.

\subsection*{Environmental Setup and Data Preparation}
Data preparation, as often is the case, can be the most time-consuming part of any data analytics project and our system development journey was not an exception to the rule. Our dataset, a single file with textual discharge summaries of 1249 patients, needed data clean-up and data staging for further data reduction. In the data clean-up step, we identified multiple abbreviations that were used to explain clinical or demographical features within our master file. While these abbreviations are useful for expediting the note taking process, they need to be translated back to full terms for the context-aware MetaMap parser to properly label them as a medical concept. For this deabbreviation, we used popular deabbreviation Perl script that was created by Solt et al. \cite{solt2009semantic}. The Perl script relies on Regular Expression (RegEx) pattern matching and replacement to deabbreviate terms back to long form. However, the script required us to first convert our text file into XML format. For this, we created a Python script to read each record and convert it to an XML document. 

The next step was to split the master file into individual patient records. We utilized Python and RegEx to search for the end of record tags and utilized that information to formulate new files for each record. Individual patient files are required by MetaMap as it tracks the position of each concept from the start of each patient record. Our end of record keyword was '[record\_end]' that facilitated boundary detection and the downstream split into new files. A master file with 1249 patient records has been split into 1249 individual patient files. 

\subsection*{Section and Boundary Detection}
Post data-preparation, our goal was to obtain a certain structure from the unstructured data. Upon visual inspection of patient documents, we observed the presence of sections within each document such as ‘PRINICIPAL DIAGNOSIS’ and ‘HISTORY OF PRESENT ILLNESS’. Based on our clinical knowledge and visual inspection of our records, we compiled a list of 15 such sections with section heading and an auto-generated unique section id. Each patient record was then parsed using string matching in Python against the compiled dictionary to detect section boundary. 

For each of 1249 patient files, we conducted string matching from the list of pre-coded sections mentioned above. Once a section heading was detected, we noted the index of the section start position (i.e. section1\_start). We continued to parse the file until we identify the starting index of a new section (i.e. section2\_start). Therefore, the section1\_end boundary was defined as section2\_start – 1. We retained all identified sections and their boundaries for each record temporarily in our Python code. 

\subsection*{Annotation Feature Extraction and Mapping}
\begin{figure}[tb]
  \centering
	\includegraphics[width=0.99\textwidth]{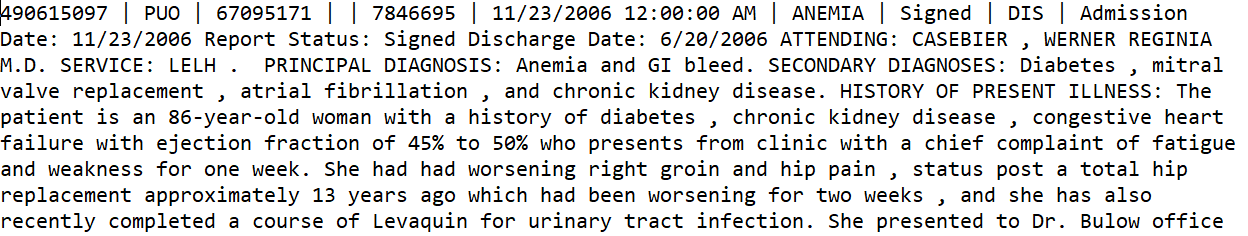}
  \caption{A snippet of the patient input file}
   \label{fig:fig1}
\end{figure}

\begin{figure}[tb]
  \centering
	\includegraphics[width=0.99\textwidth]{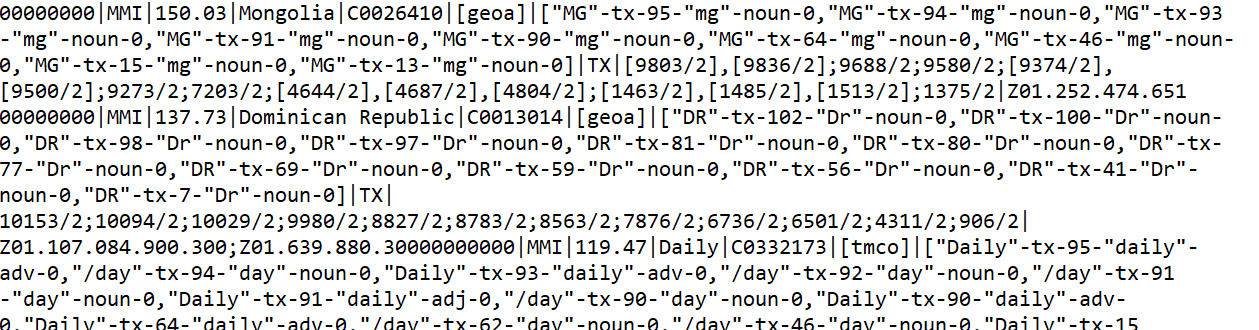}
  \caption{A snippet of MetaMap output record}
   \label{fig:fig2}
\end{figure}

MetaMap is an excellent tool that can map clinical text to the UMLS Metathesaurus concepts, which can be regarded in general as NLP (automated) annotations. MetaMap uses a knowledge-intensive approach based on symbolic, NLP and computational-linguistic techniques \cite{metamap}. Each patient file (Figure \ref{fig:fig1}) was sequentially passed through the MetaMap parser and its output was stored in individual output files (Figure \ref{fig:fig2}). We then mapped relevant MetaMap output elements to the OMOP CDM Note\_NLP Table \ref{tab:notetable}.

\begin{table}[tbp]
  \centering
  \caption{Note\_NLP table data elements}
    \begin{tabular}{|l|p{32em}|}
    \toprule
    \textbf{Column name} & \textbf{Description} \\
    \midrule
    note\_nlp\_id & A unique identifier for each term extracted from a note. A randomly generated auto-incremented number. \\
    \midrule
    note\_id & A foreign key. Thenote\_id from the Note table from the note the term was extracted from. \\
    \midrule
    section\_concept\_id & The representation of the section that extracted concept belongs to. \\
    \midrule
    snippet & A small window of the text that extracted concepts belong to. \\
    \midrule
    offset & Provided by the MetaMap in the output file. \\
    \midrule
    lexical\_variant & The actual phrase text that MetaMap generates. \\
    \midrule
    note\_nlp\_concept\_id & The concepts or CUIs. \\
    \midrule
    nlp\_system & NLP tool. \\
    \midrule
    nlp\_date\_time & Data and Time of creation/running \\
    \bottomrule
    \end{tabular}%
  \label{tab:notetable}%
\end{table}%

By utilizing the Common Data Model, we introduced standardization and portability in our system. Our system then sequentially parses each output file to load identified concepts (CUIs) including their offset (positional index) into the database. Then each loaded row, based on the offset, gets assigned to a specific section id. It is important to tag concepts to specific sections because based on the section, that concept may or may not be included as a feature for the classification.

\subsection*{Regular Expression Matches as Annotations}
Rule-based systems, and in particular systems that use regular expressions, often prove to be highly effective in tackling medical NLP problems. For example, in the i2b2 Obesity challenge, Solt et al. \cite{solt2009semantic} built a completely rule-based system that ranked first place in the intuitive task and second place in the textual task and overall first place. We value the usefulness of many existing rule-based systems and recognize the importance to introduce or improve their portability for them to be reused, adapted or extended to new corpora or phenotyping problems. This motivates us to store the key components (e.g., regular expression matches) as annotations in a common data format. For a medical record, there usually are a number of words or sentences in the record that highly suggest its category, while most of the other words or sentences are uninformative or even misleading. For example, if we capture a phrase “no evidence of coronary artery disease” from the record, it should probably be assigned as ‘Absent’ of CAD. We want to record the position of the key sentences or phrases that can help to make the classification decision. 

\begin{figure}[tb]
  \centering
	\includegraphics[width=0.99\textwidth]{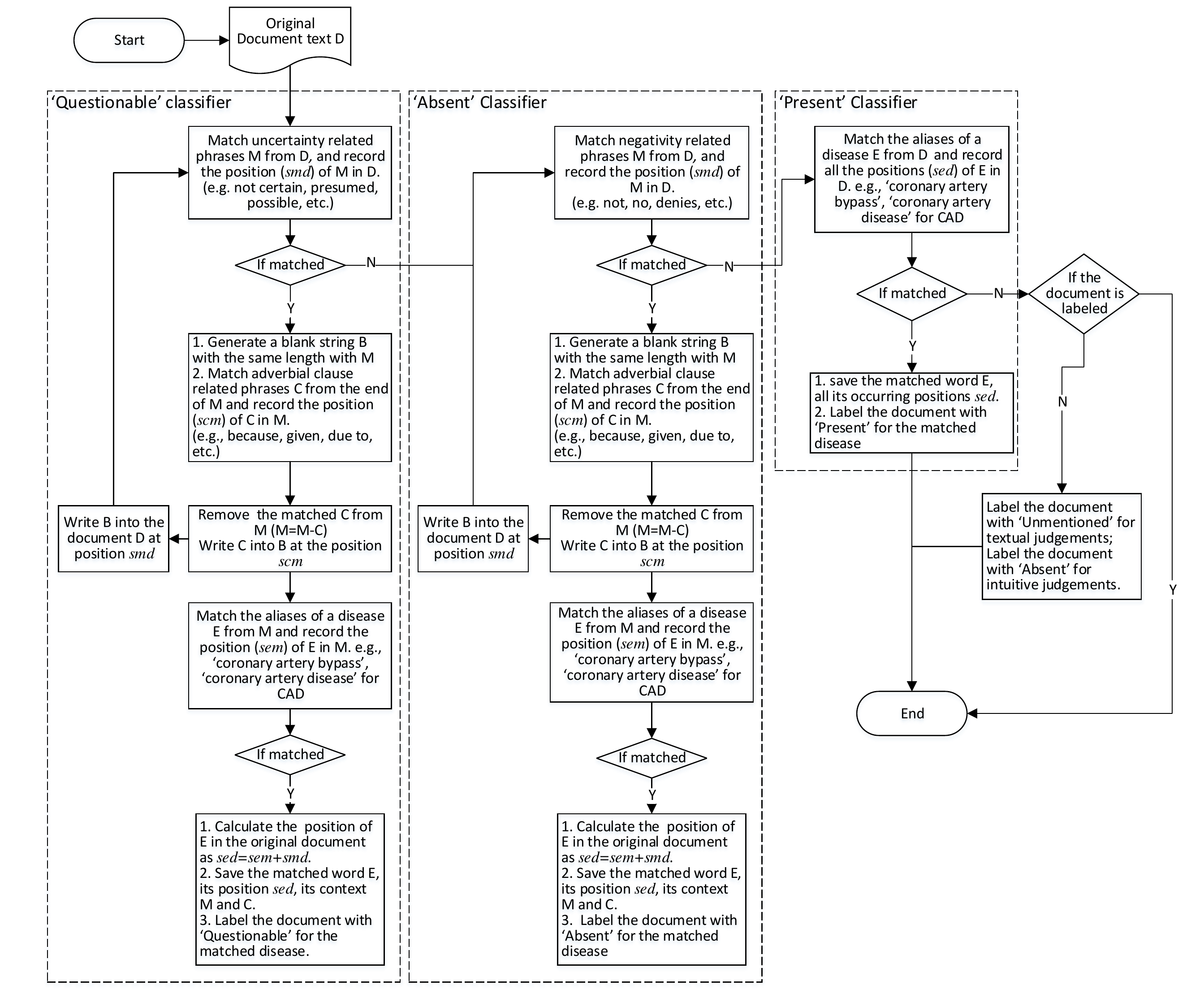}
  \caption{The word position recording process in our work}
   \label{fig:flowchart}
\end{figure}

As Solt’s rules \cite{solt2009semantic} can achieve better classification results, we follow Solt’s rule to match the category-related words or sentences. We additionally record the position of the key words or phrases when matching a RegEx, which can help to locate the key words in the original medical record. Solt's did not record the location of the word, he just removed the matched phrase from the original document for the next step match. This would change the position of the words and will make the recording of the original position difficult. For example, the Q-classifier-based rules remove the uncertainty phrases from a document before the document goes to the N-classifier for ‘Absent’ classification. Thus, when we record the position of an ‘Absent’-related word, it is no longer the position in the original record. To overcome the difficulty of recording word positions in the original document, instead of removing the matched RegEx, we replace the matched RegEx with a blank string of the same length to keep document length unchanged. Then, successive RegEx match can record the position of a word in the original text. Our word position recording process together with the document annotation process is outlined in Figure \ref{fig:flowchart}. Figure \ref{fig:flowchart} recaps the rule-based classification in Solt’s paper \cite{solt2009semantic}, and further adds our regular expression match location algorithms in order to persist the RegEx matches to OMOP CDM tables. Our design can take as input any text span. For any text span passed to the system, our algorithm will return the regular expression match position in this text span.

For each document, there can be 3 tables to save the key phrases corresponding to ‘Questionable’, ‘Absent’ and ‘Present’. For each of the tables, there are 3 fields described as follows. 
\begin{itemize}
\item disease: the name of the disease.
\item dis\_alias: the matched alias name of the disease.
\item dis\_pos: the matched position of this match in the original document (start and end position by character offset).
\end{itemize}

 For ‘Questionable’ and ‘Absent’ categories, the context of the matched disease alias is also very important. The matched RegEx should be in a sentence related to uncertainty or negation respectively. Thus, we add two more fields in the tables for words related ‘Questionable’ and ‘Absent’ to save the context of the matched RegEx. The two fields are described as follows.
 
\begin{itemize}
\item sentence: the sentence or phrase containing this match.
\item sen\_pos: the position of this sentence or phrase in the original document (start and end position by character offset).
\end{itemize}

\begin{figure}[tb]
  \centering
	\subfloat[the table for words related to ‘Questionable’]{\includegraphics[width=0.99\textwidth]{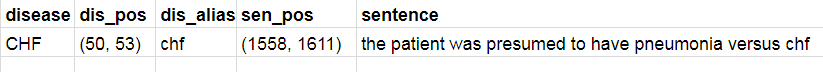}}
    
    \subfloat[the table for words related to ‘Absent’]{\includegraphics[width=0.99\textwidth]{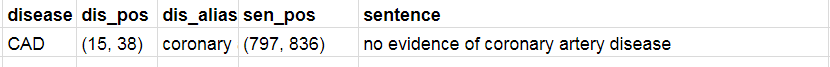}}
    
    \subfloat[ the table for words related to ‘Present’]{\includegraphics[width=0.7\textwidth]{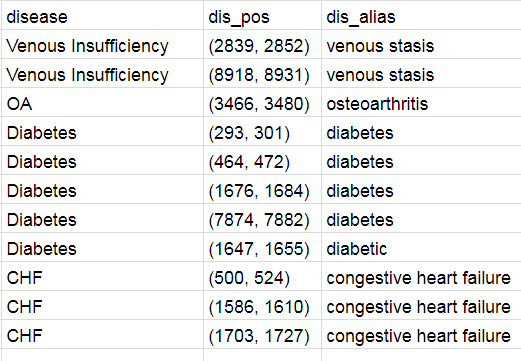}}
  \caption{ A sample of the matched regex tables.}
   \label{fig:keywordtable}
\end{figure}

Figure \ref{fig:keywordtable} shows a sample of the three tables. From these three tables, we can easily populate the OMOP CDM’s NOTE\_NLP table \ref{tab:notetable}. For example, columns offset (in the whole record) and snippet are readily computed from dis\_pos and sen\_pos. The column lexical\_variant can be populated with dis\_alias. 

\subsection*{Classification}
Since rule-based (RegEx-driven) approaches are regarded less portable between different EHR systems, we develop a machine learning based approach to improve the portability, and evaluated a range of rule-based approaches, machine learning algorithms and their mixtures to assess the trade-off between phenotyping accuracy and portability.

For each patient record, we obtain all the CUIs from the MetaMap parser. We then count the number of each CUI. This will represent the frequency of occurrence of the CUI in a medical record and serves as a feature of the record. Thus, we can construct the feature matrix based on the records and their corresponding CUIs’ frequency. We train a classification model on this feature matrix and the labels corresponding to training records and then evaluate the model using the feature matrix corresponding to the test records. In our experiment tasks, the class labels are ‘Present’, ‘Absent’ and ‘Questionable’ for intuitive judgments, and ‘Present’, ‘Absent’, ‘Questionable’ and ‘Unmentioned’ for textual judgments. To systematically evaluate the trade-off between model accuracy and portability on these data, we implement four classification methods for the classification tasks, i.e., logistic regression (LR) \cite{yu2011dual}, support vector machine (SVM) \cite{fan2008liblinear}, decision tree (DT) \cite{breiman2017classification,friedman2001elements} and random forest (RF) \cite{breiman2001random}.

\subsection*{Performance Tuning}
  For the classifiers, there are some parameters to be tuned to get better classification results. In our experiments, the parameters of the classifiers are tuned by the 3-fold cross-validated grid-search over a parameter grid \cite{hsu2003practical,chicco2017ten}. For the 4 classifiers we implemented, their parameter grids are defined in Table \ref{tab:paragrid}. For each classifier, we performed the classification for six iterations to find a better configuration for classification: a) with all CUIs, b) eliminate features from unnecessary sections, c) restrict features from clinically relevant semantic types; restrict classification to classes with statistically significant samples and then again run d) classification with all CUIs, e) eliminate features from unnecessary sections, and f) restrict features from clinically relevant semantic types.
  
\begin{table}[tbp]
  \centering
  \caption{The parameter grids for grid search}
    \begin{tabular}{|c|l|}
    \toprule
     \textbf{Classifier} & \textbf{Parameter grid} \\
    \midrule
    \textcolor[rgb]{ .114,  .122,  .133}{LR} & \textcolor[rgb]{ .114,  .122,  .133}{'C':[0.01,0.1,1,10,100]} \\
    \midrule
    \textcolor[rgb]{ .114,  .122,  .133}{SVM} & \textcolor[rgb]{ .114,  .122,  .133}{C':[0.01,0.1,1,10,100],  'kernel':['linear', 'rbf']} \\
    \midrule
    \textcolor[rgb]{ .114,  .122,  .133}{DT} & \textcolor[rgb]{ .114,  .122,  .133}{'criterion':['gini','entropy']} \\
    \midrule
    \textcolor[rgb]{ .114,  .122,  .133}{RF} & \textcolor[rgb]{ .114,  .122,  .133}{n\_estimators':[5,10,30,50,80,100],  'criterion':['gini','entropy'] } \\
    \bottomrule
    \end{tabular}%
  \label{tab:paragrid}%
\end{table}%

\section*{RESULTS AND DISCUSSION}
In our experiments, the classification performances were evaluated using micro- and macro-averaged precision (P), recall (R), and F-measure (F) \cite{uzuner2009recognizing}. Because the machine learning methods may not very effective for small sample classifications, we conducted two experiments for classification for all classes and only for the major (more populated) classes, respectively, and compared their results. In the case of classification for all classes, this setting uses standard UMLS CUI features to classify all classes for all disease phenotypes, and is considered most portable. On the contrary, entirely using Solt’s rule-based system is considered the least portable as it contains the most amount of customization (and certainly it produces the top results among challenge participants). In the middle of the spectrum there is the case of classification only for the major classes, as it integrates rule-based features using a minimal principle (where there is simply not enough training data) while retaining the standard annotation features as much as possible. Much of our results and discussions should be interpreted in the context of exposing the trade-off between portability and accuracy, as well as the parameter optimization when taking the middle-ground approach of combining rule-based features and standard UMLS CUI features. The code is available at \url{https://github.com/mocherson/portableNLP}.

\subsection*{Classification for all classes}
 Based on the above settings we obtain the classification results for all CUIs in Table \ref{tab:resallorg} (We only list the overall classification results here). From Table \ref{tab:resallorg}, we find that decision tree can achieve the best classification results among these classifiers. 
\begin{table}[tbp]
  \centering
  \caption{The classification results on all CUIs corresponding to the original records}
    \begin{tabular}{|c|c|c|c|c|c|c|}
    \toprule
    \multicolumn{7}{|c|}{intuitive} \\
    \midrule
    \multicolumn{1}{|c|}{} & \multicolumn{1}{p{4.055em}|}{P-Micro} & \multicolumn{1}{p{4.055em}|}{P-Macro} & \multicolumn{1}{p{4.055em}|}{R-Micro} & \multicolumn{1}{p{4.055em}|}{R-Macro} & \multicolumn{1}{p{4.055em}|}{F-Micro} & \multicolumn{1}{p{4.055em}|}{F-Macro} \\
    \midrule
    LR    & 0.8719 & 0.5792 & 0.8719 & 0.5509 & 0.8719 & 0.5618 \\
    \midrule
    SVM   & 0.8727 & 0.5776 & 0.8727 & 0.5537 & 0.8727 & 0.5632 \\
    \midrule
    DT    & \textbf{0.9281} & \textbf{0.6113} & \textbf{0.9281} & \textbf{0.6116} & \textbf{0.9281} & \textbf{0.6115} \\
    \midrule
    RF    & 0.8524 & 0.5626 & 0.8524 & 0.5349 & 0.8524 & 0.5454 \\
    \midrule
    \multicolumn{7}{|c|}{Textual} \\
    \midrule
    \multicolumn{1}{|c|}{} & \multicolumn{1}{p{4.055em}|}{P-Micro} & \multicolumn{1}{p{4.055em}|}{P-Macro} & \multicolumn{1}{p{4.055em}|}{R-Micro} & \multicolumn{1}{p{4.055em}|}{R-Macro} & \multicolumn{1}{p{4.055em}|}{F-Micro} & \multicolumn{1}{p{4.055em}|}{F-Macro} \\
    \midrule
    LR    & 0.8846 & 0.4379 & 0.8846 & 0.4195 & 0.8846 & 0.4268 \\
    \midrule
    SVM   & 0.8886 & 0.4384 & 0.8886 & 0.4243 & 0.8886 & 0.4300 \\
    \midrule
    DT    & \textbf{0.9436} & \textbf{0.5127} & \textbf{0.9436} & \textbf{0.5115} & \textbf{0.9436} & \textbf{0.5121} \\
    \midrule
    RF    & 0.8621 & 0.4220 & 0.8621 & 0.4044 & 0.8621 & 0.4112 \\
    \bottomrule
    \end{tabular}%
  \label{tab:resallorg}%
\end{table}%

    To disclose how a section (e.g. Family History) in the records can affect the classification results, we filter out the family history related CUIs and perform the classifications. The results are listed in Table \ref{tab:resallnofamhis}. Comparing Table \ref{tab:resallnofamhis} and Table \ref{tab:resallorg}, all the classifiers except LR can achieve higher performances without the family history than performances with it, which may indicate that family history may mislead the classification when only considering the record text for classification.
\begin{table}[tbp]
  \centering
  \caption{ The classification results without family history related CUIs}
    \begin{tabular}{|c|c|c|c|c|c|c|}
    \toprule
    \multicolumn{7}{|c|}{Intuitive} \\
    \midrule
    \multicolumn{1}{|c|}{} & \multicolumn{1}{p{4.055em}|}{P-Micro} & \multicolumn{1}{p{4.055em}|}{P-Macro} & \multicolumn{1}{p{4.055em}|}{R-Micro} & \multicolumn{1}{p{4.055em}|}{R-Macro} & \multicolumn{1}{p{4.055em}|}{F-Micro} & \multicolumn{1}{p{4.055em}|}{F-Macro} \\
    \midrule
    LR    & 0.8716 & 0.5794 & 0.8716 & 0.5503 & 0.8716 & 0.5615 \\
    \midrule
    SVM   & 0.8735 & 0.5780 & 0.8735 & 0.5546 & 0.8735 & 0.5640 \\
    \midrule
    DT    & \textbf{0.9331} & \textbf{0.6159} & \textbf{0.9331} & \textbf{0.6149} & \textbf{0.9331} & \textbf{0.6154} \\
    \midrule
    RF    & 0.8627 & 0.5685 & 0.8627 & 0.5462 & 0.8627 & 0.5551 \\
    \midrule
    \multicolumn{7}{|c|}{Textual} \\
    \midrule
    \multicolumn{1}{|c|}{} & \multicolumn{1}{p{4.055em}|}{P-Micro} & \multicolumn{1}{p{4.055em}|}{P-Macro} & \multicolumn{1}{p{4.055em}|}{R-Micro} & \multicolumn{1}{p{4.055em}|}{R-Macro} & \multicolumn{1}{p{4.055em}|}{F-Micro} & \multicolumn{1}{p{4.055em}|}{F-Macro} \\
    \midrule
    LR    & 0.8836 & 0.4372 & 0.8836 & 0.4189 & 0.8836 & 0.4262 \\
    \midrule
    SVM   & 0.8895 & 0.4391 & 0.8895 & 0.4248 & 0.8895 & 0.4306 \\
    \midrule
    DT    & \textbf{0.9475} & \textbf{0.5284} & \textbf{0.9475} & \textbf{0.5199} & \textbf{0.9475} & \textbf{0.5238} \\
    \midrule
    RF    & 0.8618 & 0.4210 & 0.8618 & 0.4049 & 0.8618 & 0.4112 \\
    \bottomrule
    \end{tabular}%
  \label{tab:resallnofamhis}%
\end{table}%

    We also conduct experiments on a list of selected CUIs without family history. We restrict our features in 15 types of CUIs which are considered most related to clinical tasks, based on clinical experiences \cite{weng2017medical} (Table \ref{tab:15cuis}). The classification results are shown in Table \ref{tab:resall15cui}. Comparing Table \ref{tab:resall15cui} and Table \ref{tab:resallnofamhis}, except for DT which can achieve the highest performances among the 4 classifiers, all other classifiers can achieve better classification performances than the performances with all CUIs. This may indicate that the 15 clinically relevant semantic types of CUIs are quite informative for classification.
\begin{table}[tbp]
  \centering
  \caption{The classification results without family history on 15 types of selected CUIs}
    \begin{tabular}{|c|c|c|c|c|c|c|}
    \toprule
    \multicolumn{7}{|c|}{Intuitive} \\
    \midrule
    \multicolumn{1}{|c|}{} & \multicolumn{1}{p{4.055em}|}{P-Micro} & \multicolumn{1}{p{4.055em}|}{P-Macro} & \multicolumn{1}{p{4.055em}|}{R-Micro} & \multicolumn{1}{p{4.055em}|}{R-Macro} & \multicolumn{1}{p{4.055em}|}{F-Micro} & \multicolumn{1}{p{4.055em}|}{F-Macro} \\
    \midrule
    LR    & 0.9024 & 0.6040 & 0.9024 & 0.5763 & 0.9024 & 0.5874 \\
    \midrule
    SVM   & 0.9077 & 0.6055 & 0.9077 & 0.5831 & 0.9077 & 0.5924 \\
    \midrule
    DT    & \textbf{0.9299} & \textbf{0.6131} & \textbf{0.9299} & \textbf{0.6129} & \textbf{0.9299} & \textbf{0.6130} \\
    \midrule
    RF    & 0.8784 & 0.5849 & 0.8784 & 0.5559 & 0.8784 & 0.5671 \\
    \midrule
    \multicolumn{7}{|c|}{Textual} \\
    \midrule
    \multicolumn{1}{|c|}{} & \multicolumn{1}{p{4.055em}|}{P-Micro} & \multicolumn{1}{p{4.055em}|}{P-Macro} & \multicolumn{1}{p{4.055em}|}{R-Micro} & \multicolumn{1}{p{4.055em}|}{R-Macro} & \multicolumn{1}{p{4.055em}|}{F-Micro} & \multicolumn{1}{p{4.055em}|}{F-Macro} \\
    \midrule
    LR    & 0.9145 & 0.4560 & 0.9145 & 0.4410 & 0.9145 & 0.4472 \\
    \midrule
    SVM   & 0.9227 & \textbf{0.5832} & 0.9227 & 0.4532 & 0.9227 & 0.4607 \\
    \midrule
    DT    & \textbf{0.9452} & 0.4878 & \textbf{0.9452} & \textbf{0.4785} & \textbf{0.9452} & \textbf{0.4807} \\
    \midrule
    RF    & 0.8830 & 0.4353 & 0.8830 & 0.4195 & 0.8830 & 0.4258 \\
    \bottomrule
    \end{tabular}%
  \label{tab:resall15cui}%
\end{table}%

\begin{table}[tbp]
  \centering
  \caption{Fifteen semantic types selected for clinical feature representations \cite{weng2017medical}}
    \begin{tabular}{|c|l|l|}
    \toprule
    \textbf{CUI} & \textbf{Semantic group} & \textbf{Semantic type description} \\
    \midrule
    T017  & Anatomy & Anatomical Structure \\
    \midrule
    T022  & Anatomy & Body System \\
    \midrule
    T023  & Anatomy & Body Part, Organ, or Organ Component \\
    \midrule
    T033  & Disorders & Finding \\
    \midrule
    T034  & Phenomena & Laboratory or Test Result \\
    \midrule
    T047  & Disorders & Disease or Syndrome \\
    \midrule
    T048  & Disorders & Mental or Behavioral Dysfunction \\
    \midrule
    T049  & Disorders & Cell or Molecular Dysfunction \\
    \midrule
    T059  & Procedures & Laboratory Procedure \\
    \midrule
    T060  & Procedures & Diagnostic Procedure \\
    \midrule
    T061  & Procedures & Therapeutic or Preventive Procedure \\
    \midrule
    T121  & Chemicals \& Drugs & Pharmacologic Substance \\
    \midrule
    T122  & Chemicals \& Drugs & Biomedical or Dental Material \\
    \midrule
    T123  & Chemicals \& Drugs & Biologically Active Substance \\
    \midrule
    T184  & Disorders & Sign or Symptom \\
    \bottomrule
    \end{tabular}%
  \label{tab:15cuis}%
\end{table}%

\subsection*{Classification for major classes}
Though machine learning based approaches are portable, compared with the total rule-based classification results listed in Table \ref{tab:ressolt}, total machine learning based classification cannot achieve good performance. Hence, we may combine rule-based approaches and machine learning algorithms to balance the classification performance and portability.
\begin{table}[t]
  \centering
  \caption{The best rule-based classification results reported in \cite{uzuner2009recognizing}}
    \begin{tabular}{|c|c|c|c|c|c|c|}
    \toprule
    \multicolumn{1}{|c|}{} & \multicolumn{1}{p{4.055em}|}{P-Micro} & \multicolumn{1}{p{4.055em}|}{P-Macro} & \multicolumn{1}{p{4.055em}|}{R-Micro} & \multicolumn{1}{p{4.055em}|}{R-Macro} & \multicolumn{1}{p{4.055em}|}{F-Micro} & \multicolumn{1}{p{4.055em}|}{F-Macro} \\
    \midrule
    Intuitive & 0.9590 & 0.7485 & 0.9590 & 0.6571 & 0.9590 & 0.6745 \\
    \midrule
    Textual & 0.9756 & 0.8318 & 0.9756 & 0.7776 & 0.9756 & 0.8000 \\
    \bottomrule
    \end{tabular}%
  \label{tab:ressolt}%
\end{table}%

 Due to the limitation of machine learning methods on small samples, in this section, we perform the classification only on the major classes that have enough samples to train a machine learning model. The class labels of the minor classes that have only a few samples are generated following Solt’s rule-based method \cite{solt2009semantic}. For intuitive judgments, we only use the ‘Present’ and ‘Absent’ records in the training data to train the classification model. For textual judgments, we only consider the ‘Present’ and ‘Unmentioned’ records. The classification results for major classes can be found in Table \ref{tab:resmajorg}, Table \ref{tab:resmajnofamhis} and Table \ref{tab:resmaj15cui} corresponding to results for all the original CUIs, all the CUIs without family history and the selected 15 types of CUIs without the family history. In Table \ref{tab:resmajorg}, Table \ref{tab:resmajnofamhis} and Table \ref{tab:resmaj15cui}, the best results are bolded, and the green shaded results can achieve the top 10 results reported in \cite{uzuner2009recognizing}.
\begin{table}[tbp]
  \centering
  \caption{The classification results for major classes on all CUIs corresponding to the original records}
    \begin{tabular}{|c|c|c|c|c|c|c|}
    \toprule
    \multicolumn{7}{|c|}{Intuitive} \\
    \midrule
    \multicolumn{1}{|c|}{} & \multicolumn{1}{p{4.055em}|}{P-Micro} & \multicolumn{1}{p{4.055em}|}{P-Macro} & \multicolumn{1}{p{4.055em}|}{R-Micro} & \multicolumn{1}{p{4.055em}|}{R-Macro} & \multicolumn{1}{p{4.055em}|}{F-Micro} & \multicolumn{1}{p{4.055em}|}{F-Macro} \\
    \midrule
    LR    & 0.8709 & \cellcolor[rgb]{ .659,  .816,  .553}0.6457 & 0.8709 & 0.5733 & 0.8709 & 0.5960 \\
    \midrule
    SVM   & 0.8724 & \cellcolor[rgb]{ .659,  .816,  .553}0.6444 & 0.8724 & 0.5770 & 0.8724 & 0.5981 \\
    \midrule
    DT    & \textbf{0.9311} & \cellcolor[rgb]{ .659,  .816,  .553}\textbf{0.6804} & \textbf{0.9311} & \cellcolor[rgb]{ .659,  .816,  .553}\textbf{0.6374} & \textbf{0.9311} & \cellcolor[rgb]{ .659,  .816,  .553}\textbf{0.6488} \\
    \midrule
    RF    & 0.8466 & 0.6226 & 0.8466 & 0.5559 & 0.8466 & 0.5765 \\
    \midrule
    \multicolumn{7}{|c|}{Textual} \\
    \midrule
    \multicolumn{1}{|c|}{} & \multicolumn{1}{p{4.055em}|}{P-Micro} & \multicolumn{1}{p{4.055em}|}{P-Macro} & \multicolumn{1}{p{4.055em}|}{R-Micro} & \multicolumn{1}{p{4.055em}|}{R-Macro} & \multicolumn{1}{p{4.055em}|}{F-Micro} & \multicolumn{1}{p{4.055em}|}{F-Macro} \\
    \midrule
    LR    & 0.8882 & \cellcolor[rgb]{ .659,  .816,  .553}0.7846 & 0.8882 & \cellcolor[rgb]{ .659,  .816,  .553}0.7085 & 0.8882 & \cellcolor[rgb]{ .659,  .816,  .553}0.7397 \\
    \midrule
    SVM   & 0.8930 & \cellcolor[rgb]{ .659,  .816,  .553}0.7858 & 0.8930 & \cellcolor[rgb]{ .659,  .816,  .553}0.7135 & 0.8930 & \cellcolor[rgb]{ .659,  .816,  .553}0.7434 \\
    \midrule
    DT    & \cellcolor[rgb]{ .659,  .816,  .553}\textbf{0.9545} & \cellcolor[rgb]{ .659,  .816,  .553}\textbf{0.8167} & \cellcolor[rgb]{ .659,  .816,  .553}\textbf{0.9545} & \cellcolor[rgb]{ .659,  .816,  .553}\textbf{0.7636} & \cellcolor[rgb]{ .659,  .816,  .553}\textbf{0.9545} & \cellcolor[rgb]{ .659,  .816,  .553}\textbf{0.7854} \\
    \midrule
    RF    & 0.8882 & \cellcolor[rgb]{ .659,  .816,  .553}0.7846 & 0.8882 & \cellcolor[rgb]{ .659,  .816,  .553}0.7085 & 0.8882 & \cellcolor[rgb]{ .659,  .816,  .553}0.7397 \\
    \bottomrule
    \end{tabular}%
  \label{tab:resmajorg}%
\end{table}%

\begin{table}[tbp]
  \centering
  \caption{The classification results for major classes without family history related CUIs}
    \begin{tabular}{|c|c|c|c|c|c|c|}
    \toprule
    \multicolumn{7}{|c|}{Intuitive} \\
    \midrule
    \multicolumn{1}{|c|}{} & \multicolumn{1}{p{4.055em}|}{P-Micro} & \multicolumn{1}{p{4.055em}|}{P-Macro} & \multicolumn{1}{p{4.055em}|}{R-Micro} & \multicolumn{1}{p{4.055em}|}{R-Macro} & \multicolumn{1}{p{4.055em}|}{F-Micro} & \multicolumn{1}{p{4.055em}|}{F-Macro} \\
    \midrule
    LR    & 0.8723 & \cellcolor[rgb]{ .659,  .816,  .553}0.6473 & 0.8723 & 0.5741 & 0.8723 & 0.597 \\
    \midrule
    SVM   & 0.8732 & \cellcolor[rgb]{ .659,  .816,  .553}0.6448 & 0.8732 & 0.5780 & 0.8732 & 0.5989 \\
    \midrule
    DT    & \textbf{0.9339} & \cellcolor[rgb]{ .659,  .816,  .553}\textbf{0.6829} & \textbf{0.9339} & \cellcolor[rgb]{ .659,  .816,  .553}\textbf{0.6392} & \textbf{0.9339} & \cellcolor[rgb]{ .659,  .816,  .553}\textbf{0.6509} \\
    \midrule
    RF    & 0.8559 & \cellcolor[rgb]{ .659,  .816,  .553}0.6317 & 0.8559 & 0.5623 & 0.8559 & 0.5838 \\
    \midrule
    \multicolumn{7}{|c|}{Textual} \\
    \midrule
    \multicolumn{1}{|c|}{} & \multicolumn{1}{p{4.055em}|}{P-Micro} & \multicolumn{1}{p{4.055em}|}{P-Macro} & \multicolumn{1}{p{4.055em}|}{R-Micro} & \multicolumn{1}{p{4.055em}|}{R-Macro} & \multicolumn{1}{p{4.055em}|}{F-Micro} & \multicolumn{1}{p{4.055em}|}{F-Macro} \\
    \midrule
    LR    & 0.8886 & \cellcolor[rgb]{ .659,  .816,  .553}0.7854 & 0.8886 & \cellcolor[rgb]{ .659,  .816,  .553}0.7083 & 0.8886 & \cellcolor[rgb]{ .659,  .816,  .553}0.7398 \\
    \midrule
    SVM   & 0.8938 & \cellcolor[rgb]{ .659,  .816,  .553}0.7865 & 0.8938 & \cellcolor[rgb]{ .659,  .816,  .553}0.7139 & 0.8938 & \cellcolor[rgb]{ .659,  .816,  .553}0.7439 \\
    \midrule
    DT    & \cellcolor[rgb]{ .659,  .816,  .553}\textbf{0.9546} & \cellcolor[rgb]{ .659,  .816,  .553}\textbf{0.8164} & \cellcolor[rgb]{ .659,  .816,  .553}\textbf{0.9546} & \cellcolor[rgb]{ .659,  .816,  .553}\textbf{0.764} & \cellcolor[rgb]{ .659,  .816,  .553}\textbf{0.9546} & \cellcolor[rgb]{ .659,  .816,  .553}\textbf{0.7855} \\
    \midrule
    RF    & 0.8640 & \cellcolor[rgb]{ .659,  .816,  .553}0.7665 & 0.8640 & \cellcolor[rgb]{ .659,  .816,  .553}0.6934 & 0.8640 & \cellcolor[rgb]{ .659,  .816,  .553}0.7233 \\
    \bottomrule
    \end{tabular}%
  \label{tab:resmajnofamhis}%
\end{table}%

\begin{table}[tbp]
  \centering
  \caption{The classification results for major classes without family history on 15 types of selected CUIs}
    \begin{tabular}{|c|c|c|c|c|c|c|}
    \toprule
    \multicolumn{7}{|c|}{Intuitive} \\
    \midrule
    \multicolumn{1}{|c|}{} & \multicolumn{1}{p{4.055em}|}{P-Micro} & \multicolumn{1}{p{4.055em}|}{P-Macro} & \multicolumn{1}{p{4.055em}|}{R-Micro} & \multicolumn{1}{p{4.055em}|}{R-Macro} & \multicolumn{1}{p{4.055em}|}{F-Micro} & \multicolumn{1}{p{4.055em}|}{F-Macro} \\
    \midrule
    LR    & 0.9001 & \cellcolor[rgb]{ .659,  .816,  .553}0.6695 & 0.9001 & 0.5979 & 0.9001 & 0.6206 \\
    \midrule
    SVM   & 0.9074 & \cellcolor[rgb]{ .659,  .816,  .553}0.6725 & 0.9074 & 0.6065 & 0.9074 & 0.6274 \\
    \midrule
    DT    & \textbf{0.9285} & \cellcolor[rgb]{ .659,  .816,  .553}\textbf{0.6783} & \textbf{0.9285} & \cellcolor[rgb]{ .659,  .816,  .553}\textbf{0.6355} & \textbf{0.9285} & \cellcolor[rgb]{ .659,  .816,  .553}\textbf{0.6467} \\
    \midrule
    RF    & 0.8690 & \cellcolor[rgb]{ .659,  .816,  .553}0.6417 & 0.8690 & 0.5740 & 0.8690 & 0.5952 \\
    \midrule
    \multicolumn{7}{|c|}{Textual} \\
    \midrule
    \multicolumn{1}{|c|}{} & \multicolumn{1}{p{4.055em}|}{P-Micro} & \multicolumn{1}{p{4.055em}|}{P-Macro} & \multicolumn{1}{p{4.055em}|}{R-Micro} & \multicolumn{1}{p{4.055em}|}{R-Macro} & \multicolumn{1}{p{4.055em}|}{F-Micro} & \multicolumn{1}{p{4.055em}|}{F-Macro} \\
    \midrule
    LR    & 0.9188 & \cellcolor[rgb]{ .659,  .816,  .553}0.8037 & 0.9188 & \cellcolor[rgb]{ .659,  .816,  .553}0.7303 & 0.9188 & \cellcolor[rgb]{ .659,  .816,  .553}0.7608 \\
    \midrule
    SVM   & 0.9273 & \cellcolor[rgb]{ .659,  .816,  .553}0.806 & 0.9273 & \cellcolor[rgb]{ .659,  .816,  .553}0.7388 & 0.9273 & \cellcolor[rgb]{ .659,  .816,  .553}0.7669 \\
    \midrule
    DT    & \cellcolor[rgb]{ .659,  .816,  .553}\textbf{0.9538} & \cellcolor[rgb]{ .659,  .816,  .553}\textbf{0.8160} & \cellcolor[rgb]{ .659,  .816,  .553}\textbf{0.9538} & \cellcolor[rgb]{ .659,  .816,  .553}\textbf{0.7633} & \cellcolor[rgb]{ .659,  .816,  .553}\textbf{0.9538} & \cellcolor[rgb]{ .659,  .816,  .553}\textbf{0.7849} \\
    \midrule
    RF    & 0.8864 & \cellcolor[rgb]{ .659,  .816,  .553}0.7823 & 0.8864 & \cellcolor[rgb]{ .659,  .816,  .553}0.7081 & 0.8864 & \cellcolor[rgb]{ .659,  .816,  .553}0.7386 \\
    \bottomrule
    \end{tabular}%
  \label{tab:resmaj15cui}%
\end{table}%
 
    From Table \ref{tab:resmajorg}, Table \ref{tab:resmajnofamhis} and Table \ref{tab:resmaj15cui}, we can draw a consistent conclusion with previous analysis that the Family History section may mislead the classification and the 15 clinically relevant semantic types of CUIs can be useful for these classifiers except DT. In addition, by combining the rule-based approach and machine learning based approaches, we can achieve a comparable classification performance with the total rule-based approach, and more importantly, this method can be portable between different HER systems.by combining the rule-based approach and machine learning based approaches, we can achieve a comparable classification performance with the total rule-based approach, and more importantly, this method can be portable between different EHR systems. This is as expected due to the limitation of machine learning methods on small samples. Thus, in our portable phenotyping system, we can use the rule-based method for the minor class classification and use machine learning methods for the major class classification.

\section*{CONCLUSION}
 Recently, increasing amount of patient data is becoming electronically available. To handle the explosion of EHR data, healthcare professionals and researchers will increasingly rely on automated or semi-automated computational techniques to derive knowledge from these data. Significant effort has been devoted to the implementation of open-sourced, standard-based systems to improve the portability of electronic health record (EHR)-based phenotype definitions (e.g., eMERGE \cite{gottesman2013electronic} and PhEMA \cite{rasmussen2015modular}). We developed a portable phenotyping system that is capable of integrating both rule-based and statistical machine learning based phenotyping approaches. Our system can mine and store both standard UMLS features and the key features of rule-based systems (e.g., regular expression matches) from the unstructured text as NLP annotations using the format defined by the OMOP CDM, in order to standardize necessary data elements. Comparing to file system based pipelines such as UIMA CAS stacks and BioC, the OMOP CDM uses a database as the persistent storage and has the advantages offered by database management systems. This includes well-defined schemas, remote queries and query optimizations. We demonstrated that we can store NLP annotations including those from concepts from standard pipelines (e.g., MetaMap), regular expression matches, and section annotations in CDM tables, which can later be used for computational phenotyping. Our system can thus enable the development of new standard UMLS feature-based NLP systems as well as the reuse, adaptation and extension of many existing rule-based clinical NLP systems. Given the highly variable nature of unstructured biomedical data and evolving machine learning techniques, future researchers may also benefit by adopting a similar iterative approach to optimizing their classification and by using mixed classification methods. However, variation in data models and coding systems used at different institutions make it difficult to conduct a large-scale analysis of observational healthcare databases. Our system is a first step to address that problem and enhances its portability by utilizing the OMOP CDM and its standardized terminologies. Once data (raw input and processed output) from multiple sources get harmonized into the Common Data Model, researchers can conduct systematic analysis at larger scale to perfect these new secondary research techniques in biomedical data mining, Natural Language Processing, Machine Learning etc. By breaking down the barriers of institutional variability with portable systems and standardized terminologies, we can unlock the hidden potential in our biomedical and health data. We note that we have not explored how the CDM can be applied to tasks other than phenotyping/classification tasks and will leave it as future work to explore how CDM can lend value to other types of tasks as well.

%


\begin{backmatter}



\section*{Acknowledgements}
  We would like to thank i2b2 National Center for Biomedical Computing funded by U54LM008748, for providing the clinical records originally prepared for the Shared Tasks for Challenges in NLP for Clinical Data organized by Dr. Ozlem Uzuner. We thank Dr. Uzuner for helpful discussions. This work was supported in part by NIH Grants 2R01GM105688-06 and 1R21LM012618-01.
  

\bibliographystyle{bmc-mathphys} 
\bibliography{bmc_article}      

\end{backmatter}
\end{document}